\documentclass[10pt, a4paper]{article}

\usepackage{lrec-coling2024} 
\usepackage{amsmath}

\usepackage{graphicx}
\usepackage{breqn}
\usepackage{nopageno}
\usepackage{times}
\usepackage{latexsym}
\usepackage{hyperref}
\usepackage{array}

\usepackage[T1]{fontenc}

\usepackage[utf8]{inputenc}

\usepackage{microtype}

\title{\textbf{STEntConv: Predicting Disagreement with Stance Detection and a Signed Graph Convolutional Network}}

\name{Isabelle Lorge $^{1}$, Li Zhang  $^{1,2}$ \thanks{Work completed while Li Zhang was a postdoctoral assistant at the University of Oxford.}, Xiaowen Dong$^{1}$, Janet B. Pierrehumbert$^{1}$} 

\address{$^{1}$ Department of Engineering, University of Oxford\\
$^{2}$ Institute of Finance and Technology, University College London \\
isabelle.lorge@psych.ox.ac.uk, ucels07@ucl.ac.uk, \\ xiaowen.dong@eng.ox.ac.uk, janet.pierrehumbert@oerc.ox.ac.uk}

\abstract{
The rise of social media platforms has led to an increase in polarised online discussions, especially on political and socio-cultural topics such as elections and climate change. We propose a simple and novel unsupervised method to predict whether the authors of two posts agree or disagree, leveraging user stances about named entities obtained from their posts. We present STEntConv, a model which builds a graph of users and named entities weighted by stance and trains a Signed Graph Convolutional Network (SGCN) to detect disagreement between comment and reply posts. We run experiments and ablation studies and show that including this information improves disagreement detection performance on a dataset of Reddit posts for a range of controversial subreddit topics, without the need for platform-specific features or user history.
 \\ \newline  }

\begin{document}

\maketitleabstract

\section{Introduction}
Social media now form an integral part of many people's lives. While these tools have allowed users unprecedented access to shared content, ideas and views across the world, they have also permitted the fast rise and spread of harmful forms of communication, such as fake news, abuse and communities acting as radicalising echo chambers at unseen scales \cite{terren2021echo}. It is then of high interest to investigate the polarisation of opinions as a reflection of ever-shifting political and socio-cultural dynamics which have direct impact on society. For example, detecting disagreement between users can help assess the controversiality of a topic, give insight into user opinions which would not be obtainable from their post in isolation or provide a way to estimate numbers for sides of a debate. 

\begin{figure}[h]
    \centering
    \includegraphics[scale=0.30]{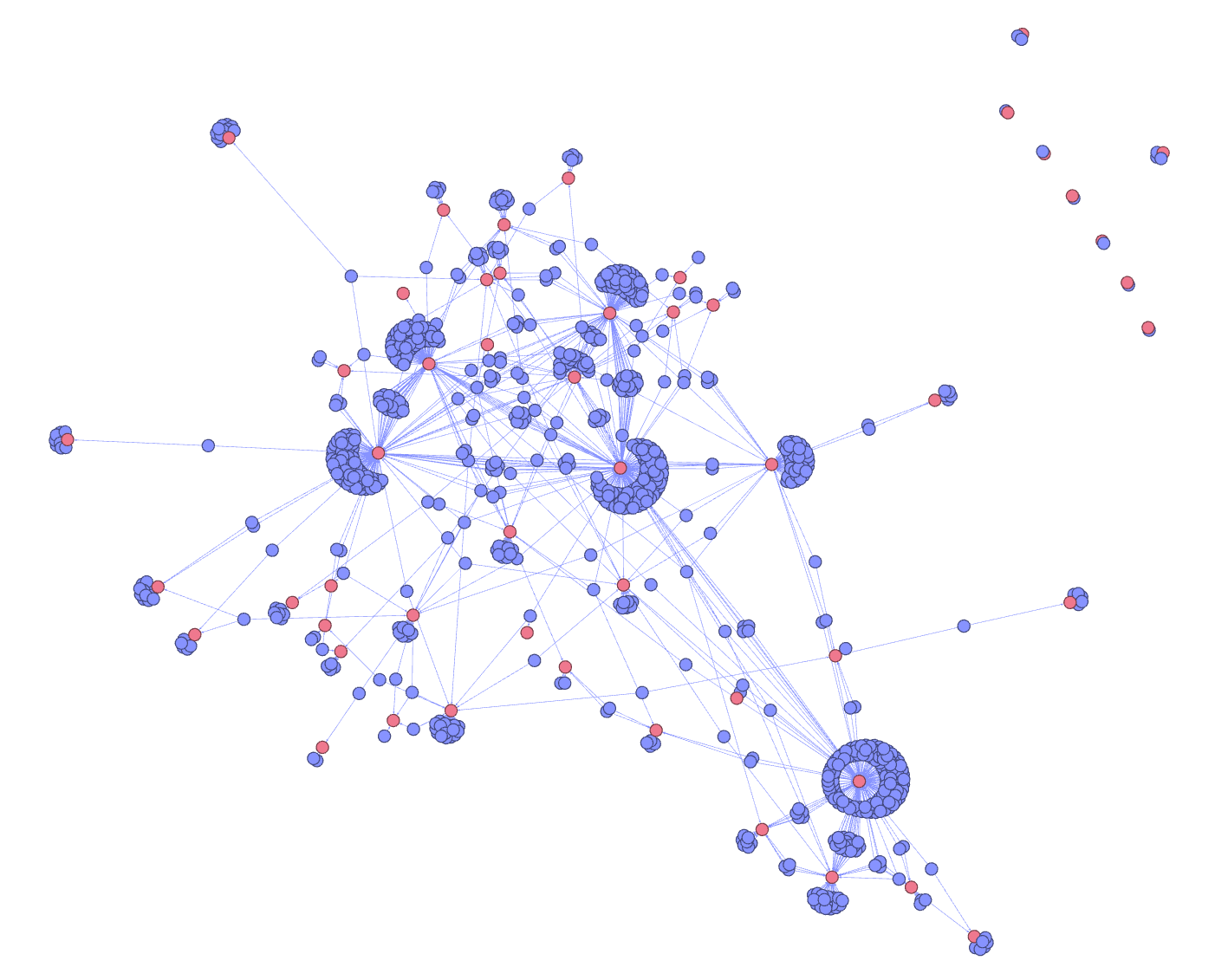}
    \caption{User-entity graph visualised with Gephi \cite{bastian2009gephi} (positive edges). We apply a force atlas layout. Pink nodes are entities, blue nodes are users which we can see clustered around the target entities they expressed a positive stance for.}
    \label{fig:entities_graph}
\end{figure}

Online communities constitute an ideal terrain for this investigation, as they are likely to foster various tensions and debates and allow researchers to examine them in real time or longitudinally  \cite{alkhalifa2022capturing}. 
Previous work on detecting disagreement has focused on supplementing textual information with user network information, either gathered through platform-specific features such as Twitter's following system, retweets and hashtags (which cannot be generalised across platforms) (e.g., \citealp{darwish2020unsupervised} or through user-user interaction history (which is not necessarily available) (e.g., \citealp{luo2023improving}). 
Instead, to the best of our knowledge we are the first to represent users through a user-entity signed graph weighted by stance. In addition to being generalisable to any platform and not requiring user interaction history, our method has the potential to provide more explainable representations for users by explicitly tracing disagreements back to entities they feel positively or negatively about. Furthermore, the graph can easily be adapted to various controversial topics by selecting entities relevant to that topic, is able to accommodate different amounts of information per user, and a user-entity signed network constitutes a natural and explicit representation of polarising allegiances (cf. figure\ref{fig:entities_graph}). Finally, we derive stance towards entities in an unsupervised manner which means there is no need to obtain costly manual labels. 

We make the choice to use BERT both for unsupervised stance detection and for the textual part of the model itself. This choice is partially to be able to directly assess the additional contribution of our signed network to the former best model from \citet{pougue2021debagreement} for this dataset, but also because we build it with the view of a relatively lightweight model with potential for real time applications. In addition, large language models have been shown to underperform smaller state-of-the-art fine-tuned models on specific tasks, for example in the biomedical domain (\citealp{ateia2023chatgpt}). However, we do provide results for the performance of an open-source large language model (Falcon, \citealt{falcon40b}) on the task as a comparison.

Our main contributions are as follows\footnote{We make all our code and data available at \url{https://github.com/isabellelorge/contradiction}}:

1) We offer a simple, unsupervised method to extract user stances towards entities by leveraging sentence-BERT.

2) We build a model using a weighted Signed Graph Convolutional Network on a user-entity graph with BERT embeddings to detect disagreement, improving on previous state-of-the-art results on a dataset of Reddit posts.

3) We present various model ablation studies and demonstrate the robustness of the proposed framework.

We start by outlining current research regarding stance and signed graphs which is relevant to our task. We then move on to describing the dataset used and graph extracted from our data, the architecture and various parameters of our model. We finally present experimental results and discuss their implications.

\section{Background}
\subsection{Stance}
The word stance refers to the intellectual or emotional attitude or position of an author towards a specific concept or entity, such as atheism or the legalisation of abortion \cite{mohammad2016semeval}. This is different from the concept of sentiment as it is usually defined in sentiment analysis, where it refers to the overall emotion expressed by a piece of text. A given text can then have one sentiment value but express multiple positive and negative stances, the target of which is not necessarily explicitly mentioned in the text.

Some concepts lend themselves more easily to the elicitation of stance. For example, consider the following quote from the Wikipedia section on Donald Trump:
\begin{quote}
\textbf{Donald John Trump} (born \textbf{June 14, 1946}) is an \textbf{American} politician, media personality, and businessman who served as the \textbf{45th} president of \textbf{the United States} from \textbf{2017 to 2021}. \textbf{Trump}'s political positions have been described as populist, protectionist, isolationist, and nationalist. He won the \textbf{2016} presidential election as the \textbf{Republican} nominee against \textbf{Democratic} nominee \textbf{Hillary Clinton} despite losing the popular vote.
\end{quote}

It would be strange to ask whether the author is in favour or against \textit{born, media, businessman, who, is, from,  described, presidential} or \textit{popular}. On the other hand, the words in bold type \footnote{Bold typed words are all word spans identified by Spacy \cite{spacy2} as named entities.} (with the exception of numerals, i.e. dates and ordinals) seem like more appropriate targets for holding an opinion. These words are generally referred to as \textit{Named Entities} or NEs, a term first coined for the Sixth Message Understanding Conference (MUC-6) \cite{grishman-sundheim-1996-message}. The category aims to encompass expressions which are \textit{rigid designators}, as defined by \citet{Kripke1980-KRINAN}, i.e., which designate the same object in all possible worlds in which that object exists and never designate anything else. In other words, these expressions refer to specific instances in the world, including but not limited to proper names. Common NE categories are: organisations, people, locations (including states), events, products and quantities (including dates, times, percent, money, quantity, ordinals and cardinals). 

With the exception of the last category, most of these constitute valid targets for extracting author stance because, unlike other terms (e.g., verbs, adverbs, prepositions, some nouns and adjectives) they can be involved in debates and elicit diverging intellectual or emotional viewpoints. Communities tend to be created on the basis of shared traits which are either given (e.g., race, gender, nationality) or acquired (preferences and opinions). Thus, for many contentious issues, agreement and disagreement between individuals are likely to crystallise around attitudes towards a few key entities which define membership identity in a form of `neo-tribalism' \cite{maffesoli1995time}. 
Stance can be modelled at the post or author level. Here we choose to leverage all posts from a known author to determine their stance toward a specific entity.

\subsection{Signed Graphs}
Graphs, defined as combinations of nodes and edges, are useful abstractions for a variety of structures and phenomena. They can take several forms: directed (e.g., Twitter following) vs. undirected (e.g., Facebook friends); signed (e.g., likes and dislikes) vs. unsigned (e.g., retweets), homogenous vs. bipartite (with nodes of different types where there is no between-type edges, e.g., employees and companies they worked for).
In the current paper, given we model user-entity stances, the graph  constructed is a signed bipartite graph. While it is technically directed (the stance is from user towards entity), there are no edges in the opposite direction (i.e., from entity to user), thus we treat the graph as undirected for simplicity. 

Various methods have been developed for node representation in graphs. When no node features are available, methods relying on connectivity and random walks such as DeepWalk \cite{perozzi2014deepwalk} and node2vec \cite{grover2016node2vec} can produce low-dimensional node embeddings using a similar algorithm to skip-gram in Word2vec, i.e., by predicting a node given previously encountered nodes. Graph Neural Networks (GNNs), on the other hand, can leverage both connectivity and node features from a local neighbourhood to produce node representations. Among these, Graph Convolutional Networks (GCNs) were first introduced by \citet{kipf2016semi} and constitute a popular option akin to a generalisation of Convolutional Neural Networks (CNNs), by performing a first-order approximation of a spectral filter on a neighbourhood. 



GCNs were originally designed to handle unsigned graphs. However, in the case of stance, as well as in many other applications related to social media, we encounter networks which are signed, i.e., which involve positive and negative edges. Processing these types of graphs and producing meaningful node representations is not straightforward, as there is an intrinsic qualitative difference between the two types of edges which cannot be optimally resolved by e.g., treating them alike, ignoring negative edges, or ignoring edges that cancel each other.  One solution is to keep positive and negative representations separate from the graph neural network separate and simply concatenate them. Another way suggested by \citet{DBLP:journals/corr/abs-1808-06354} relies on assumptions from \textit{balance theory} \cite{heider1946attitudes, cartwright1956structural}, which comes from social psychology and formalises intuitions such as `an enemy of an enemy is a friend'. Thus, for each layer $l$, the aggregation function would gather on the positive side not only friendly nodes, but friends of friends and enemies of enemies, and similarly on the negative side get information from enemies but also friends of enemies and enemies of friends. The positive and negative convolutions are then concatenated together to produce the final node representations as in the simpler model. In our experiments we test both the simple signed model and the model with additional aggregations based on balance theory.

\section{Dataset}

\begin{table*}
\centering
\begin{tabular}{ c c c c c c } 
 \hline
  & \textit{r/Brexit} & \textit{r/climate}  & \textit{r/BlackLivesMatter}  & \textit{r/Republican} & \textit{r/democrats}  \\ 
 \hline
 \textbf{start date} & 
 Jun 2016  &
 Jan 2015 &
 Jan 2020 &
 Jan 2020 &
 Jan 2020 \\ 
 \hline
 \textbf{agree} & 0.29 & 0.32 & 0.45 & 0.34 & 0.42 \\
 
 \textbf{neutral} & 0.29 & 0.28 & 0.22 & 0.25 & 0.22 \\
 
 \textbf{disagree} & 0.42 & 0.40  & 0.33  & 0.41 & 0.36 \\
 \hline
\end{tabular}
\caption{DEBAGREEMENT statistics per subreddit and period}
\label{dataset}
\end{table*}

\begin{table*}
\centering
\begin{tabular}{c c c c} 
\hline
& \textit{comment-reply count} & \textit{avg length (comment)}  & \textit{avg length (reply)}  \\
\hline
\textit{r/Brexit} & 15745 & 45  & 40\\

\textit{r/climate} & 5773 & 43  & 41\\

\textit{r/BlackLivesMatter} & 1929 &  41 & 39 \\

\textit{r/Republican} & 9823 &  38 & 35 \\

\textit{r/Democrats} & 9624 &  38 & 37 \\
\hline
\end{tabular}
\caption{DEBAGREEMENT post counts and word lengths }
\label{dataset_full}
\end{table*}

We use the \textit{DEBAGREEMENT} dataset for our experiments \cite{pougue2021debagreement}, a dataset of 42894 Reddit comment-reply pairs from 5 different subreddits (\textit{r/Brexit, r/climate, r/BLM, r/Republican} and \textit{r/democrats}) with each pair given one of three labels: \textit{agree/neutral/disagree} (see dataset statistics in tables \ref{dataset} and \ref{dataset_full}). The pairs of posts were labelled by crowdsourcers who received intensive training on the issues discussed in the various subreddits. The disagreement prediction task consists in predicting which of the three labels describes the relation between the comment and reply posts.

This is a very difficult task for a number of reasons. First, assessing disagreement around issues such as those discussed in the selected subreddits requires expert knowledge (hence the need for specific training of crowdsourcers). Second, there is a high level of subjectivity involved, which is evidenced  by the `clean' version of the dataset still containing over 60\% labels where only 2 out of 3 crowdsourcers agreed. Because of the latter, we choose after examining the data to work with the portion of the dataset that was given the same label by all three crowdsourcers (16723 pairs of posts). Finally, it is worth noting that most previous works focusing on disagreement use Twitter data exclusively (e.g., \citealp{darwish2020unsupervised, trabelsi2018unsupervised, zhou2023stance}). Many other platforms like Reddit lack network features such as common hashtags, user following and retweets which are highly useful for detecting endorsement between users. It is then much harder to create user representations indicative of polarisation. This emphasises the crucial need to find alternative features, such as user-entity stances, which can generalise across platforms. While the task has been tackled using user network features \cite{luo2023improving}, no previous works have attempted to improve performance without leveraging user interaction features which may not always be available.

\section{Framework}

\begin{table*}
\centering
\begin{tabular}{c} 
 \hline
     \textit{american, antifa, aoc, asian, backstop, bernie, biden, black, blm, brexit,} \\
     \textit{brexiteers, brown, christian, cnn, communist, con, confederate, conservative,}\\ 
     \textit{corbyn, cuomo, dem, democrat, democratic, dems, dnc, fascist, fbi, floyd, george, gop,} \\
     \textit{greta, holocaust, jew, kkk, leave, leftist, liberal, libertarian, maga, marxist,}\\
     \textit{mcconnell, moderate, moron, msm, muslim, nazi, party, patriot, pete, poc,} \\
     \textit{progressive, propaganda, qanon, racist, referendum, remainers, republican, riot,} \\
     \textit{romney, sander, senate, statue, tory, trump, tucker, warren, white} \\ 
 \hline
\end{tabular}
\caption{Extracted target entities}
\label{entities}
\end{table*}

\begin{table*}
\centering
\begin{tabular}{c c c c c c c c c c} 
 \hline
& \textit{$|\mathcal{U}|$} & \textit{$|\mathcal{A}|$}  & \textit{$|\mathcal{E+}|$} & \textit{$|\mathcal{E-}|$}  &
\textit{$|\mathcal{D}|$} &
\textit{$|\mathcal{D(U)}|$} &
\textit{$|\mathcal{D(A)}|$} &
\textit{$|\mathcal{CN(U)}|$} &
\textit{$|\mathcal{CN(A)}|$} 
\\ 
 \hline
 \hline
 \textit{train} & 7107  & 67  & 3997 & 4615 & 0.001 &
 1.83 & 194  & 0.32  & 5.67 \\
  \textit{test} &  1513 & 67   & 863  & 866  & 0.002  &  1.48  & 37   & 0.20 & 0.60  \\
 \hline
\end{tabular}
\caption{User-entity graph statistics for full training and test datasets. \textit{$|\mathcal{U}|$}: number of users; \textit{$|\mathcal{A}|$}: number of entities;  \textit{$|\mathcal{E+}|$}: number of positive edges;
\textit{$|\mathcal{E-}|$}: number of negative edges ; 
\textit{$|\mathcal{D}|$}: graph density; 
\textit{$|\mathcal{D(U)}|$}: average degree (users); 
\textit{$|\mathcal{D(A)}|$}: average degree(entities);
\textit{$|\mathcal{CN(U)}|$}: average common neighbors (users); \textit{$|\mathcal{CN(A)}|$}: average common neighbors (entities)}

\label{graph}
\end{table*}

\subsection{User-Entity Graph Construction}

\begin{figure}[h]
    \centering
    \includegraphics[scale=0.35]{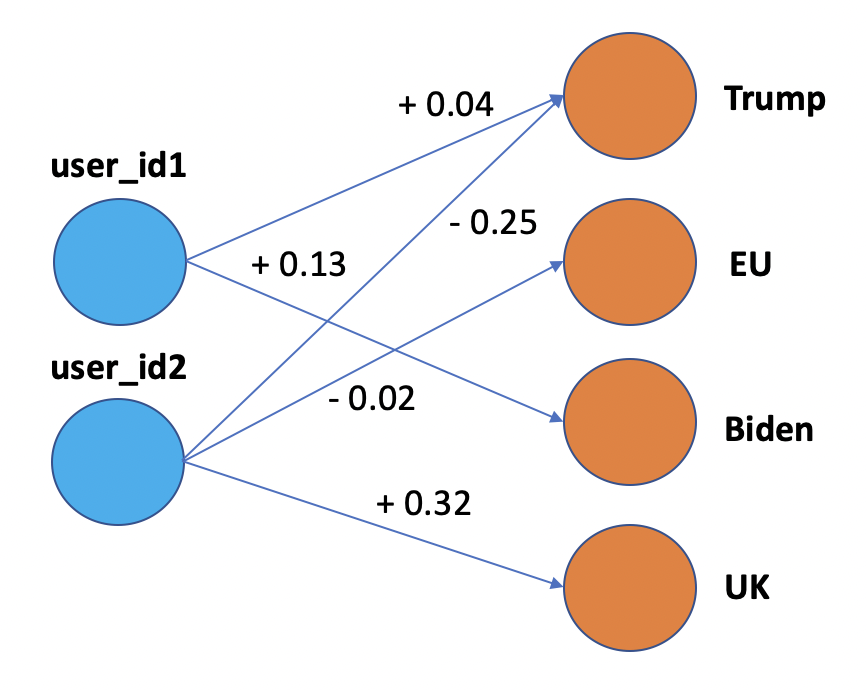}
    \caption{Example user-entity graph. The network is signed, with each edge representing user stance towards an entity.}
    \label{fig:graph}
\end{figure}

Let $\mathcal{G} = (\mathcal{N}, \mathcal{E})$ be a signed undirected bipartite graph where $\mathcal{U} \in \mathcal{N}$ is the set of user nodes, $\mathcal{A} \in \mathcal{N}$ is the set of entity nodes and $\mathcal{E}$ the set of edges between users and entities, with $\mathcal{E+}$ the set of positive edges and $\mathcal{E-}$ the set of negative edges. Since this is a bipartite graph, there are no edges between users or between entities, and the set of positive and negative edges are defined to be mutually exclusive (ie., there is at most one edge, either positive or negative, between a user and an entity) (see figure \ref{fig:graph}). 

We build the graph in the following way. First, we extract named entities for each comment and reply post using Spacy \cite{spacy2}, discarding entities which pertain to the categories `CARDINAL', `DATE', `ORDINAL', `WORK\_OF\_ART', `PERCENT', `QUANTITY' and `MONEY'. Since we do not have ground truth for the stance of each author for each extracted entity, we devise an unsupervised method to obtain a proxy of it by leveraging Sentence-BERT \cite{reimers-2019-sentence-bert}. For each entity, we create `pro' and `con' sentences using the templates \textit{I am for X} and \textit{I am against X}. We then compute the cosine similarity between each SBERT-embedded post sentence and each SBERT-embedded template sentence and subtract the `con' cosine similarity from the `pro' cosine similarity \footnote{While this method has to our knowledge not been previously used, we manually examine the results for 100 sentences by calculating the stance for each named entity extracted using the method described and rating it as correct or incorrect and find satisfactory performance (0.68 accuracy).}. Finally, we take the mean of all cosine differences for an entity across all user posts\footnote{In addition, we also mean center all edge weight values by subtracting the mean.}. The advantage of this method is that it is almost entirely unsupervised and does not require prior domain knowledge or manual selection of relevant topics or entities (only the subreddit titles, see below), these naturally arise among the most frequent entities extracted from the corpus. Therefore, we define our stance measure as:

\begin{equation}
stance_{u, e} = \sum^{N}_{i\in P}\frac{\sum^{M}_{s\in S} \frac{cos_{s}^{+} -  cos_{s}^{-}}{|S|}} {|P|}
\end{equation}

\noindent where $stance_{u, e}$ is the stance of user $u$ towards entity $e$, $i \in P$ is the ith post contributed by user $u$, $s \in S$ is a sentence in the post, $cos^{+}$ is the cosine similarity of the post sentence with the `pro' embedded template sentence and $cos^{-}$ is the cosine similarity of the post sentence with the `con' embedded template sentence. We notice a negative bias in our extracted cosine similarities whereby the mean $\mu$ of stance values lies around -0.02 and accordingly split edges into positive edges ($stance_{u,e} >= \mu$) and negative edges ($stance_{u,e} < \mu$) after verifying that the stances follow a normal distribution and the median is close to the mean. The statistics of the resulting graph can be seen in table \ref{graph}.

When examining the extracted entities, it appears that most entities which occur only a few times in the corpus will be irrelevant to our task. We thus apply a combination of two filters: we keep entities which are amongst the 5000 most frequent entities and whose embeddings have a cosine similarity above 0.5 to at least one subreddit title embedding (\textit{Brexit, climate, BLM,
Republican} and \textit{democrats}) (the embeddings used are the initial features for the GCN which are Word2vec embeddings trained on our dataset, see section Training).
We obtain both values by conducting a sensitivity correlation analysis on the training set in the following way: we model a negative and positive entity vector for each author respectively as the sum of the negative stance and positive stance entities, concatenate them and measure the Kendall $\tau$ rank correlation between the cosine similarity of the author vectors for a given comments pair and the label (0,1 or 2 as disagreement, neutral and agreement). There is a clear peak in correlation with entities which have over 0.5 cosine similarity with at least one subreddit title and are within the 5000 most frequent entities, thus we select these as our threshold values.
We also filter out multiword entities which often show redundancy and misextractions. The final set of 67 target entities can be seen in Table \ref{entities}, a heatmap of cosine similarities to each subreddit can be found in Appendix \ref{sec:appendixA} and a visualisation of the user-entity graph can be seen in figure \ref{fig:entities_graph}.

To get a fair assessment of our model and be able to directly compare it with the performance of the GCN model alone, we subset the training dataset to comment-reply pairs which mention at least one of our target entities (for other comment-reply pairs the GCN would not have any features). The final dataset is made of 1770 comment-reply pairs. While this constitutes only 10\% of the original full agreement dataset, we notice that disagreements which are most closely related to the subreddit's controversial topic will often contain the target entities
\footnote{While our assumption that the model leverages information from entities not present in the text suggests we should be able to use posts which do not mention target entities, we find empirically that this is not the case. However, the better performance over the BERT baseline suggests the model does make use of information not present in the comment-reply pair. We hypothesise that this is because of a correlation between the presence of target entities in specific pairs of posts and the amount of additional information in the entity graph (ie., posts which do not contain target entities tend to come from authors for whom there is little/no entity information). }
. We also believe that given the difficulty of the task (especially on Reddit data where many network features available on Twitter cannot be used), an improvement on a subset of disagreement types is worthwhile and holds promise for applicability. We do provide results for the subset of the dataset where only either comment or reply mentions one of our target entities, which constitutes about 40\% of the full agreement dataset or 6174 comment-reply pairs, however we cannot run a comparison with the GCN model alone in this case.

\subsection{STEntConv}

\begin{figure}
    \flushleft
    \includegraphics[scale=0.5]{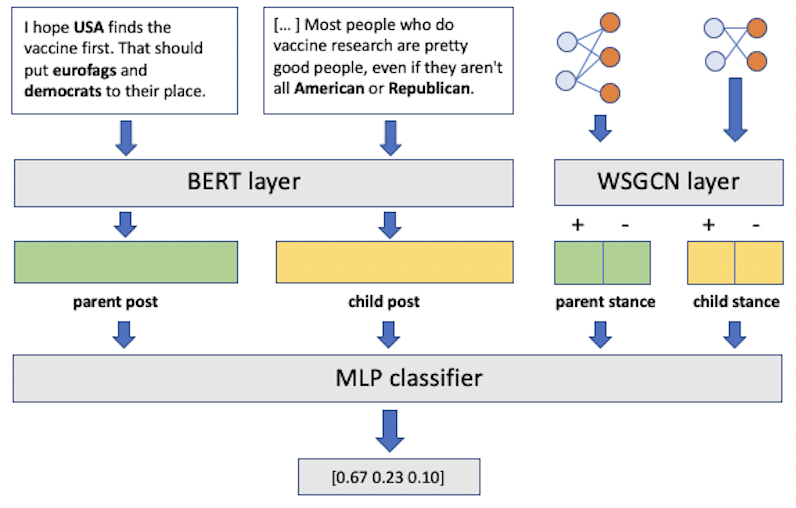}
    \caption{Model architecture.}
    \label{fig:model}
\end{figure}

We 
adopt the Signed Graph Convolutional Network proposed in \citet{DBLP:journals/corr/abs-1808-06354} and modify it to integrate edge weights for our stance values so that the positive and negative convolutions are as follows:

{\small
\begin{equation}
\begin{split}
\textbf{h}^{B(l)}_{i} = \sigma( \textbf{W}^{B(l)} [ 
\sum_{j\in \mathcal{N}_{i}^{+}}
\frac{\textbf{h}^{B(l-1)}_{j}}{|\mathcal{N}^{+}_{i}|}
\textbf{w}_{j},  \\
\sum_{k\in \mathcal{N}_{i}^{-}}
\frac{\textbf{h}^{U(l-1)}_{k}}{|\mathcal{N}^{-}_{i}|}
\textbf{w}_{k},
\textbf{h}^{B(l-1)}_{i}
]),
\end{split}
\end{equation}

\begin{equation}
\begin{split}
\textbf{h}^{U(l)}_{i} = \sigma( \textbf{W}^{U(l)} [ 
\sum_{j \in\mathcal{N}_{i}^{+}}
\frac{\textbf{h}^{U(l-1)}_{j}}
{|\mathcal{N}^{+}_{i}|}
\textbf{w}_{j}, \\
\sum_{k\in \mathcal{N}_{i}^{-}}
\frac{\textbf{h}^{B(l-1)}_{k}}{|\mathcal{N}^{-}_{i}|}
\textbf{w}_{k},
\textbf{h}^{U(l-1)}_{i}
]),
\end{split}
\end{equation}}

\noindent where $\textbf{h}^{B(l)}_{i}$ is the weighted aggregation of positive edges for layer $l$, $\sum_{j \in\mathcal{N}_{i}^{+}}
\frac{\textbf{h}^{B(l-1)}_{j}}
{|\mathcal{N}^{+}_{i}|}
\textbf{w}_{j}$ is the weighted sum of `friends of friends', $\sum_{k\in \mathcal{N}_{i}^{-}}
\frac{\textbf{h}^{U(l-1)}_{k}}{|\mathcal{N}^{-}_{i}|}
\textbf{w}_{k}$ is the weighted sum of `enemies of enemies' and 
$\textbf{h}^{B(l-1)}_{i}$ the previous layer's positive edges aggregation. Similarly, $\textbf{h}^{U(l)}_{i}$ is the aggregation of negative edges for layer $l$,  $\sum_{j \in\mathcal{N}_{i}^{+}}
\frac{\textbf{h}^{U(l-1)}_{j}}
{|\mathcal{N}^{+}_{i}|}
\textbf{w}_{j}$ is the weighted sum of `enemies of friends', $\sum_{k\in \mathcal{N}_{i}^{-}}
\frac{\textbf{h}^{B(l-1)}_{k}}{|\mathcal{N}^{-}_{i}|}
\textbf{w}_{k}$ is the weighted sum of `friends of enemies' and 
$\textbf{h}^{U(l-1)}_{i}$ the previous layer's negative edges aggregation. We run experiments both with the additional aggregations from \textit{balance theory} and without (i.e., only aggregating direct friends for positive edges and direct enemies for negative edges), in which case the respective weighted aggregations are simply:
\begin{equation}
\textbf{h}^{B(l)}_{i} = \sigma( \textbf{W}^{B(l)} [ 
\sum_{j\in \mathcal{N}_{i}^{+}}
\frac{\textbf{h}^{(l-1)}_{j}}{|\mathcal{N}^{+}_{i}|}
\textbf{w}_{j},
\textbf{h}^{(l-1)}_{i}
]).
\end{equation}
\begin{equation}
\textbf{h}^{U(l)}_{i} = \sigma( \textbf{W}^{U(l)} [ 
\sum_{j\in \mathcal{N}_{i}^{+}}
\frac{\textbf{h}^{(l-1)}_{j}}{|\mathcal{N}^{-}_{i}|}
\textbf{w}_{j},
\textbf{h}^{(l-1)}_{i}
]),
\end{equation} 
This is also the definition of the aggregations for the first layer $l$ = 1. We build the weighted version of the algorithm by adapting the unweighted \citet{DBLP:journals/corr/abs-1808-06354} implementation from PyTorch geometric (PyG) \cite{Fey/Lenssen/2019}. The rationale for integrating edge weights to the convolutional layer is that, given our unsupervised method for calculating stance, a high absolute value is more reliable and thus considered more informative. Positive/negative edge and node features should then be weighted accordingly when performing message passing (e.g., a small edge weight is more likely to denote a stance close to neutral). The output of the GCN is concatenated to the output of a BERT layer for comment and reply posts and fed to a one-layer feed-forward network. The final model architecture can be seen in figure \ref{fig:model}.

\section{Baselines}
\subsubsection{BERT}
As a baseline, we fine-tune a BERT (base, uncased) layer as was originally used by \citet{pougue2021debagreement}, i.e., we ablate the graph convolutional layer from our model and feed this to the linear layer for classification.
\subsubsection{GCN only}
In addition, we conduct the opposite ablation, i.e., we use only the GCN to assess how the model performs relying only on the positive and negative edges of the stance graph without any access to the text of the posts.
\subsubsection{StanceRel}
\citet{luo2023improving} improve on previous results on the DEBAGREEMENT dataset by building a graph autoencoder and training it on a signed undirected user-user interaction graph which creates user representations based on their previous interactions (agreement or disagreement, i.e., positive or negative) and using the decoded user features along with textual features to detect disagreement. 
While a direct comparison with our approach may not be fully informative (as the two models use different features which may be available in different situations), we train their model on our subset of data and provide results on our test set to give a picture of the differential value of various user features. 
\subsubsection{FALCON} We also test an instruct-trained version of Falcon \cite{falcon40b}, specifically \textit{vilsonrodrigues/falcon-7b-instruct-sharded} implemented through the transformers library \cite{DBLP:journals/corr/abs-1910-03771}. Prompt and hyperparameters used can be found in Appendix \ref{sec:appendixB}.

\section{Training}
We train 100-dimensional word2vec\footnote{We also experiment with GloVe embeddings but the performance is worse.} embeddings  on the full dataset and use the resulting vectors as initial features for our entities. User features are initialised as 100-dimensional vectors of zeros. We obtain contextual text embeddings for comment and reply posts through the \textit{transformers} library \cite{DBLP:journals/corr/abs-1910-03771} implementation of a BERT (base, uncased) layer whose output we mean pool, excluding special tokens, and concatenate these together with the output of our weighted Signed Graph Convolutional Network layer before feeding this to a one-layer linear classifier. Since the classes are not entirely balanced, we compute class weights and weigh class loss accordingly during training. We use cross entropy as loss function, a batch size of 16, a hidden size of 300 for the first convolutional layer, a learning rate of 3e-5 and the Adam optimiser with weight decay 1e-5. We split the data into 0.80 train, 0.10 dev and 0.10 test and train for 6 epochs (models with BERT layers) and 11 epochs (GCN only). We experiment with number of convolutional layers (one versus two), type of aggregation (balance theory or only direct friends and enemies), edge weights (binary versus weighted) and sentences used to calculate stance (full post versus only sentences containing target entity). We train the models with three different random seeds and average the results.


\section{Results}

\begin{table*}
\centering
\begin{tabular}{p{4cm} c c c c c c} 
 \hline
   & \textit{r/Brexit} & \textit{r/climate}  & \textit{r/Republican}  & \textit{r/democrat} & \textit{r/BLM*} & \textbf{all (sd)}  \\ 
    \hline

  \textit{(c|r) \textbf{BERT}}  &  .75 & \textbf{.79}  & .73  & .69  & .72   & .72 (0.03)\\
 
  \textit{(c|r) \textbf{STEntConv}} 
  & \textbf{.78} & .78   & \textbf{.76}   & \textbf{.71}   & \textbf{.75}  & \textbf{.75} (0.02)\\
 \hline
     \textit{(c\&r) \textbf{FALCON}}  & .40  & .25  & .45  & .38   & 1.0   & .42  (0.28) \\
  \textit{(c\&r) \textbf{BERT}}  & .58  & .54 & .69 & .63  & .67  &  .64 (0.06)\\
    \textit{(c\&r) \textbf{StanceRel}}  & .67  & .30  & .67  & .60   & 1.0   & .65  (0.22) \\

  \textit{(c\&r) \textbf{STEntConv} (GCN)} & .36 & .44  & .44  & .37  & .67 & .43 (0.11)\\
  
  \textit{(c\&r) \textbf{STEntConv} (m.agg)} & \textbf{.70} & .41  & \textbf{.73}  & .69  & 1.0  & .70  (0.18) \\
 \textit{(c\&r) \textbf{STEntConv}} & .62  & \textbf{.64 } & .70  & \textbf{.74}  & \textbf{1.0}  & \textbf{.71} (0.14) \\

 \hline
\end{tabular}
\caption{Macro averaged F1 for each model and subreddit. \textbf{STEntConv} = our model enhanced with entity stances; \textbf{BERT}= BERT model (base, uncased); \textbf{StanceRel} = relation graph model from \citet{luo2023improving} \textbf{FALCON}: Falcon model (instruct trained, 7B); \textit{GCN} = STEntConv without BERT component; \textit{m.agg}: multiple aggregations, i.e. using the `friend of friend' additional aggregation from \citet{DBLP:journals/corr/abs-1808-06354}. (c\&r) = dataset with target entity in comment and reply; (c|r) = dataset with target entity in comment or reply. Best in bold.*The (c\&r) test set only contained one comment-reply pair from the \textit{r/BLM} subreddit. 
}
\label{results}
\end{table*}
Results can be found in Table \ref{results}. The best performing model uses one convolution layer, only direct friends and enemies, weighted edges, and the full text of the post for stance extraction. As can be seen, on the \textit{(c\&r)} subset the addition of the user stance graph information helps improve model performance by 7 points on average compared to the BERT baseline and 6 points over the StanceRel model which previously obtained the best results on this task. While the improvement in performance is weaker for the version of STEntConv trained on the \textit{(c|r)} dataset (since this dataset includes authors for whom the model has no relevant stance information), the model still achieves a 3 point increase  over the BERT baseline. 

In the non-multiple aggregation model, the boost from the stance graph information is lowest for \textit{r/Republican}. This is consistent with additional analyses we conduct on the test dataset showing that the \textit{r/Republican} subreddit has the highest ratio of target entities present in posts versus all entity information available for authors, meaning that there is little extra information from the GCN to be used and most relevant information is already available to the BERT model. Thus, it is likely that the better performance of our model is due to STEnTConv being able to leverage stance information about entities not present in the  comment-reply pair being classified. 

\begin{figure}[h]
    \centering
    \includegraphics[scale=0.35]{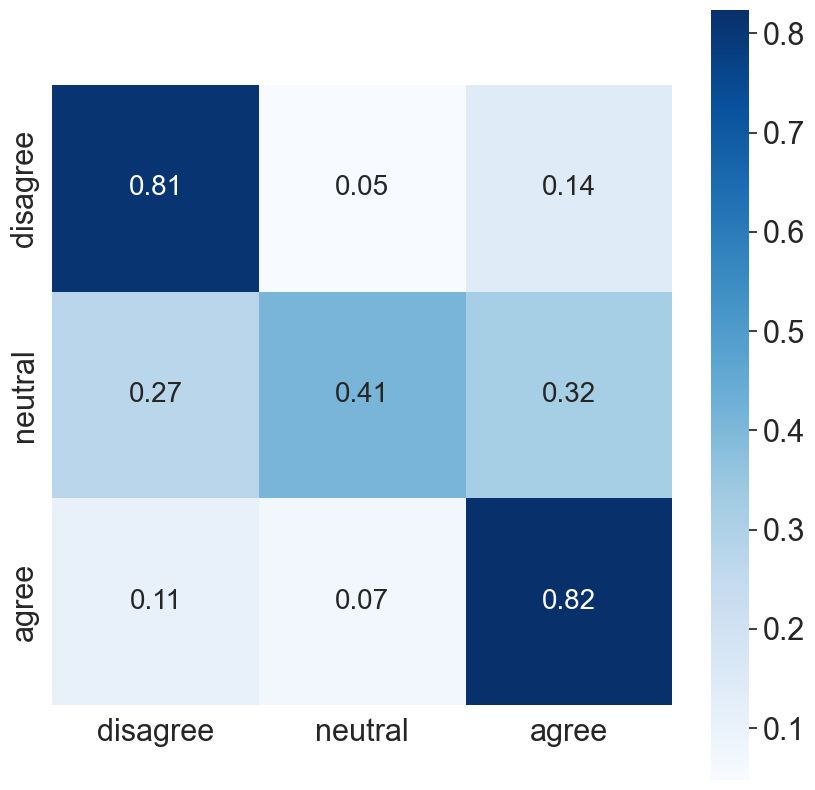}
    \caption{Confusion matrix (STEntConv)}
    \label{fig:confusion_matrix}
\end{figure}

Adding the second aggregation from \cite{DBLP:journals/corr/abs-1808-06354} performs better on the \textit{r/Brexit} and \textit{r/Republicans} subreddits. This would tend to indicate that the additional aggregations for these subreddits remain relevant to the task whereas they introduce noise for \textit{r/climate}. This is supported by \textit{r/climate} subreddit having the lowest cosine similarity on average to the target entities. Falcon performs particularly poorly, in addition to requiring over an hour for inference on the test set (vs. a few seconds for BERT-based models and GCN). StanceRel, while above the BERT baseline, underperforms our model on the test set. Given that the model leverages user history which represents a strong additional signal, we expected it to perform better. The lower performance may be due to sparseness in the user interaction graph or to the model requiring a larger dataset to learn features from interaction history. In addition, the graph StanceRel uses is weighted by frequency which may be prone to biases (e.g., user A from the Green Party agreed with user B from the Green Party many times, but they disagreed on an issue involving a specific entity, e.g. French nuclear company AREVA). Finally, they leverage the assumptions from balance theory which we show may not be useful for this particular task. Unsurprisingly, the neutral class is hardest to classify, with an f1 of 0.4 versus 0.8 for the agree/disagree classes. Examining the data indicates one reason for this might be the heterogenous nature of the class, with some neutral pairs of posts discussing unrelated topics and others agreeing and disagreeing in equal amounts. The confusion matrix between classes can be seen in figure \ref{fig:confusion_matrix}.

\section{Related Work}

There have been a number of studies aimed at modelling stance and disagreement. Regarding stance, a supervised task was introduced at SemEval 2016 along with a small dataset of 4870 tweets annotated for stance against five topics (\textit{Atheism, Climate Change is a Real Concern, Feminist Movement, Hillary Clinton} and \textit{Legalization of Abortion}) \cite{mohammad2016semeval}. Winning teams used CNNs and ensemble models. However, the limited data and specificity of both style and topics make it difficult to extend a model trained on this data to detecting stance in other domains. Regarding unsupervised methods for both stance and disagreement, most methods have focused on modelling users through dimensionality reduction and clustering of content \cite{darwish2020unsupervised} or through platform-specific \cite{trabelsi2018unsupervised, zhou2023stance} or non platform-specific \cite{luo2023improving} network features from user-user interactions. To the best of our knowledge, no previous method used a user-entity graph to model user representations.

Importantly, as stated in the introduction, most previous works use Twitter data which contains platform-specific network features. Thus, our work aligns with other efforts to build alternative network features 
 in polarised communities when user endorsement features (e.g., retweets and follows) are not available, as is the case for Reddit data. For example, \citet{hofmann-etal-2022-modeling} build a graph of edges between social entities (concepts and subreddits) to identify the level of polarisation for a given concept. In addition, even previous works which do not use platform-specific data still rely on features which may not always be available, such as user interaction history \cite{luo2023improving} and restrict ability to use graph features to known users. By contrast, our method can leverage entity stances at inference time to model novel users which have similar allegiances to users seen in training. To our knowledge, the current paper is the first to use representations from a signed graph between users and entities for the purpose of predicting disagreement.

\section{Conclusion}
We presented a simple, unsupervised and domain-agnostic pipeline for creating graph user features to improve disagreement detection between comment-reply pairs of social media posts. We ran several experiments against baselines and performed ablations to examine the contribution of model components and parameters. Our model uses GCN convolutions over a signed bipartite graph of automatically extracted user-entity stances. STEntConv can be leveraged to create comprehensive user representations which take into account stance towards various target entities in order to better predict whether two users are likely to agree or disagree on a range of controversial topics, regardless of availablity of platform-specific network features or user interaction history. As a next step, this method could easily be extended to target entities beyond named entities to include common nouns which are particularly relevant to the controversial topic, especially in cases where the topic is less likely to involve named entities.  
\section{Limitations}
\textbf{Scale.}
We acknowledge that the improvement we demonstrate over the baseline applies to only a subset of the original dataset. However, given the difficulty of the task and the lack of additional network features for Reddit data we believe this improvement is still worthwhile. Furthermore, our method could potentially be extended to include stance towards relevant common nouns in addition to named entities.

\noindent\textbf{Domain.}
While the dataset we used covers a range of controversial topics from socio-cultural to political issues, it was extracted from 5 specific subreddits and thus it is uncertain to what extent our results would apply to other topics of disagreement and whether the model could be generalised to other domains. However, we note that our pipeline for extracting relevant entities is entirely domain-agnostic and thus we believe it could be applied successfully to any forum debating  a controversial topic.

\section*{Acknowledgements}
The authors would like to acknowledge support from the EPSRC (EP/T023333/1). X.D. acknowledges support from the Oxford-Man Institute of Quantitative Finance. This study was supported by EPSRC Grant
EP/W037211/1.


\section{References}\label{sec:reference}

\bibliography{custom}
\bibliographystyle{lrec-coling2024-natbib}

\onecolumn
\newpage
\appendix

\section{Subreddit-entity heatmap}

\label{sec:appendixA}

\begin{figure}[h]
    \centering
    \includegraphics[scale=0.50]{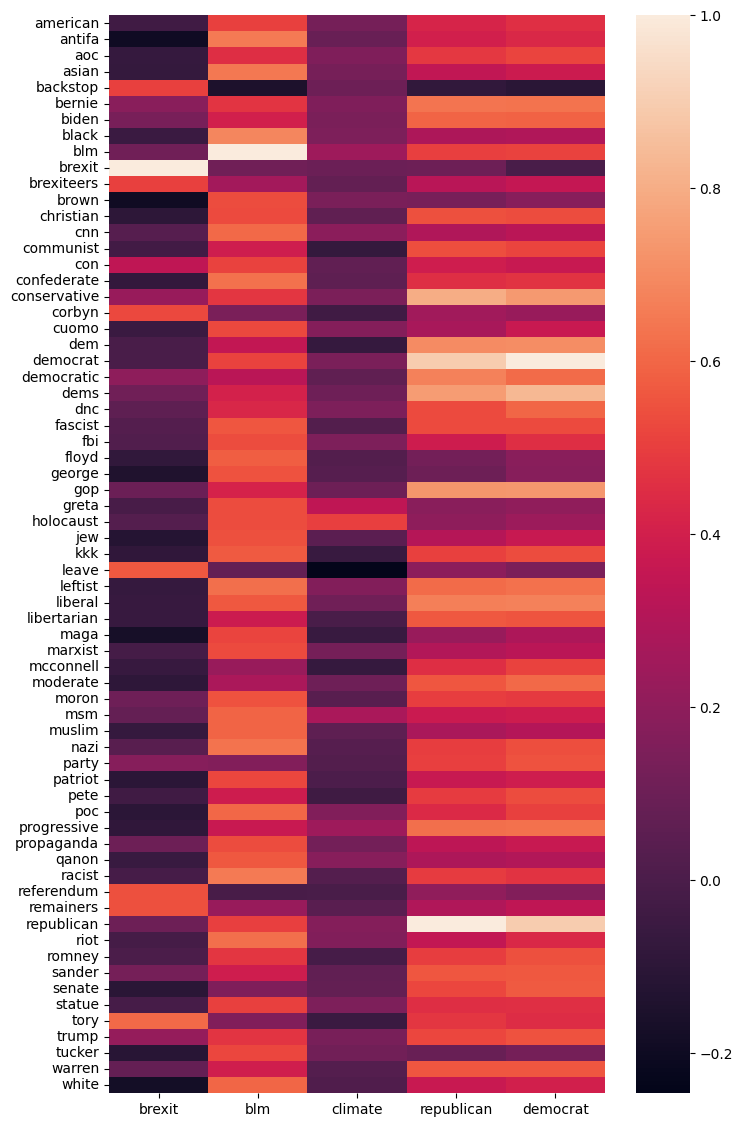}
    \caption{heatmap of cosine similarities between each subreddit name and target named entities.}
    \label{fig:heatmap}
\end{figure}

\section{Falcon prompt and hyperparameters}
\label{sec:appendixB}
\begin{verbatim}


quantization_config = BitsAndBytesConfig(load_in_4bit=True,
    bnb_4bit_compute_dtype=torch.float16,
    bnb_4bit_quant_type="nf4",
    bnb_4bit_use_double_quant=True)


model_id = "vilsonrodrigues/falcon-7b-instruct-sharded"


model_4bit = AutoModelForCausalLM.from_pretrained(model_id,
        device_map="auto",
        quantization_config=quantization_config,
        trust_remote_code=True)


tokenizer=AutoTokenizer.from_pretrained(model_id)


pipeline=transformers.pipeline("text-generation",
          model=model_4bit,
          tokenizer=tokenizer,
          use_cache=True,
         device_map="auto",
          max_length=38 + len(comment)+ len(reply),
         do_sample=False,
          top_k=10,
          num_return_sequences=1,
         eos_token_id=tokenizer.eos_token_id,
          pad_token_id=tokenizer.eos_token_id)
          

  prompt = f```
    Here are a social media COMMENT and a REPLY. 
    
    Say whether the reply is agreeing, disagreeing or neutral towards the comment: 
    
    COMMENT: {comment} 
    
    REPLY: {reply} 
    
    The reply is
    '''
\end{verbatim}



\end{document}